\documentclass{article}

\usepackage[nonatbib, final]{neurips_2019}

\usepackage[utf8]{inputenc} 
\usepackage[T1]{fontenc}    
\usepackage{hyperref}       
\usepackage{url}            
\usepackage{booktabs}       
\usepackage{amsfonts}       
\usepackage{nicefrac}       
\usepackage{microtype}      

\title{Short Text Language Identification for Under Resourced Languages}

\author{%
  Bernardt Duvenhage\\
  Feersum Engine\\
  Praekelt Consulting\\
  Johannesburg, South Africa \\
  \texttt{bernardt@praekelt.com} \\
}

\begin{document}

\maketitle

\begin{abstract}
The paper presents a hierarchical naive Bayesian and lexicon based classifier for short text language identification (LID) useful for under resourced languages. The algorithm is evaluated on short pieces of text for the 11 official South African languages some of which are similar languages. 

The algorithm is compared to recent approaches using test sets from previous works on South African languages as well as the Discriminating between Similar Languages (DSL) shared tasks' datasets. Remaining research opportunities and pressing concerns in evaluating and comparing LID approaches are also discussed. 
\end{abstract}


\section{Introduction}
Accurate language identification (LID) is the first step in many natural language processing and machine comprehension pipelines. If the language of a piece of text is known then the appropriate downstream models like parts of speech taggers and language models can be applied as required. 

LID is further also an important step in harvesting scarce language resources. Harvested data can be used to bootstrap more accurate LID models and in doing so continually improve the quality of the harvested data. Availability of data is still one of the big roadblocks for applying data driven approaches like supervised machine learning in developing countries.

Having 11 official languages of South Africa has lead to initiatives (discussed in the next section) that have had positive effect on the availability of language resources for research. However, many of the South African languages are still under resourced from the point of view of building data driven models for machine comprehension and process automation.

Table~\ref{tab:languages} shows the percentages of first language speakers for each of the official languages of South Africa. These are four conjunctively written Nguni languages (zul, xho, nbl, ssw), Afrikaans (afr) and English (eng), three disjunctively written Sotho languages (nso, sot, tsn), as well as tshiVenda (ven) and Xitsonga (tso). The Nguni languages are similar to each other and harder to distinguish. The same is true of the Sotho languages. 

This paper presents a hierarchical naive Bayesian and lexicon based classifier for LID of short pieces of text of 15-20 characters long. The algorithm is evaluated against recent approaches using existing test sets from previous works on South African languages as well as the Discriminating between Similar Languages (DSL) 2015 and 2017 shared tasks. \footnote{Code and datasets available at \url{https://github.com/praekelt/feersum-lid-shared-task}.} 

Section~\ref{sec:related_works} reviews existing works on the topic and summarises the remaining research problems. Section~\ref{sec:methodology} of the paper discusses the proposed algorithm and Section~\ref{sec:results} presents comparative results.

\begin{table}
  \caption{Percentage of South Africans by First Language}
  \label{tab:languages}
  \centering
  \begin{tabular}{llll}
    \toprule
    Language (ISO 639) & Percentage & Language (ISO 639) & Percentage \\
    \midrule
    IsiZulu (zul) & 23\% & isiNdebele (nbl) & 2.1\% \\
    IsiXhosa (xho) & 16\% & siSwati (ssw) & 2.5\% \\
    \midrule
    Afrikaans (afr) & 14\% & English (eng) & 9.6\% \\
    \midrule
    Sepedi (nso) & 9.1\% & Setswana (tsn) & 8.0\% \\
    Sesotho (sot) & 7.6\% \\
    \midrule
    Xitsonga (tso) & 4.5\% & Tshivenda (ven) & 2.4\% \\
    \bottomrule
  \end{tabular}
\end{table}

\section{Related Works}
\label{sec:related_works}
The focus of this section is on recently published datasets and LID research applicable to the South African context. An in depth survey of algorithms, features, datasets, shared tasks and evaluation methods may be found in \cite{Jauhiainen2018}.

The datasets for the DSL~2015 \& DSL~2017 shared tasks~\cite{Zampieri2017} are often used in LID benchmarks and also available on Kaggle~\footnote{https://www.kaggle.com/vardial/dslcc}. The DSL datasets, like other LID datasets, consists of text sentences labelled by language. The 2017 dataset, for example, contains 14 languages over 6 language groups with 18000 training samples and 1000 testing samples per language. 

The recently published JW300 parallel corpus~\cite{Agic2019} covers over 300 languages with around 100 thousand parallel sentences per language pair on average. In South Africa, a multilingual corpus of academic texts produced by university students with different mother tongues is being developed \cite{Carstens2019}. The WiLI-2018 benchmark dataset~\cite{Thoma2018} for monolingual written natural language identification includes around 1000 paragraphs of 235 languages. A possibly useful link can also be made~\cite{FrancoSalvador2017} between Native Language Identification (NLI) (determining the native language of the author of a text) and Language Variety Identification (LVI) (classification of different varieties of a single language) which opens up more datasets. The Leipzig Corpora Collection~\cite{Goldhahn2012}, the Universal Declaration of Human Rights~\footnote{\url{http://research.ics.aalto.fi/cog/data/udhr/}} and Tatoeba~\footnote{\url{https://tatoeba.org/eng/downloads}} are also often used sources of data. 

The NCHLT text corpora~\cite{NCHLTTextCorpora2014} is likely a good starting point for a shared LID task dataset for the South African languages~\cite{Duvenhage2017}. The NCHLT text corpora contains enough data to have 3500 training samples and 600 testing samples of 300+ character sentences per language. Researchers have recently started applying existing algorithms for tasks like neural machine translation in earnest to such South African language datasets~\cite{Abbott2018}.




Existing NLP datasets, models and services~\cite{Puttkammer2018} are available for South African languages. These include an LID algorithm~\cite{Hocking2014} that uses a character level n-gram language model. Multiple papers have shown that 'shallow' naive Bayes classifiers~\cite{EspichnLinares2018, Duvenhage2017, Barbaresi2017, Jauhiainen2017}, SVMs~\cite{Medvedeva2017} and similar models work very well for doing LID. The DSL~2017 paper~\cite{Zampieri2017}, for example, gives an overview of the solutions of all of the teams that competed on the shared task and the winning approach~\cite{Bestgen2017} used an SVM with character n-gram, parts of speech tag features and some other engineered features. The winning approach for DSL~2015 used an ensemble naive Bayes classifier. The fasttext classifier~\cite{Joulin2017} is perhaps one of the best known efficient 'shallow' text classifiers that have been used for LID~\footnote{\url{https://fasttext.cc/blog/2017/10/02/blog-post.html}}. 

Multiple papers have proposed hierarchical stacked classifiers (including lexicons) that would for example first classify a piece of text by language group and then by exact language~\cite{Goutte2014, Zampieri2015,Duvenhage2017, Jauhiainen2018}. Some work has also been done on classifying surnames between Tshivenda, Xitsonga and Sepedi~\cite{Sefara2016}. Additionally, data augmentation~\cite{Marivate2019} and adversarial training~\cite{Li2018} approaches are potentially very useful to reduce the requirement for data.

Researchers have investigated deeper LID models like bidirectional recurrent neural networks~\cite{Kocmi2017} or ensembles of recurrent neural networks~\cite{Mathur2017}. The latter is reported to achieve 95.12\% in the DSL~2015 shared task. In these models text features can include character and word n-grams as well as informative character and word-level features learnt~\cite{Jaech2016} from the training data. The neural methods seem to work well in tasks where more training data is available.

In summary, LID of short texts, informal styles and similar languages remains a difficult problem which is actively being researched. Increased confusion can in general be expected between shorter pieces of text and languages that are more closely related. Shallow methods still seem to work well compared to deeper models for LID. Other remaining research opportunities seem to be data harvesting, building standardised datasets and creating shared tasks for South Africa and Africa. Support for language codes that include more languages seems to be growing and discoverability of research is improving with more survey papers coming out. Paywalls also seem to no longer be a problem; the references used in this paper was either openly published or available as preprint papers.

\section{Methodology}
\label{sec:methodology}
The proposed LID algorithm builds on the work in \cite{Duvenhage2017} and \cite{Malmasi2015}. We apply a naive Bayesian classifier with character (2, 4 \& 6)-grams, word unigram and word bigram features with a hierarchical lexicon based classifier. 

The naive Bayesian classifier is trained to predict the specific language label of a piece of text, but used to first classify text as belonging to either the Nguni family, the Sotho family, English, Afrikaans, Xitsonga or Tshivenda. The scikit-learn multinomial naive Bayes classifier is used for the implementation with an alpha smoothing value of 0.01 and hashed text features.

The lexicon based classifier is then used to predict the specific language within a language group. For the South African languages this is done for the Nguni and Sotho groups. If the lexicon prediction of the specific language has high confidence then its result is used as the final label else the naive Bayesian classifier's specific language prediction is used as the final result. The lexicon is built over all the data and therefore includes the vocabulary from both the training and testing sets.

The lexicon based classifier is designed to trade higher precision for lower recall. The proposed implementation is considered confident if the number of words from the winning language is at least one more than the number of words considered to be from the language scored in second place. 

The stacked classifier is tested against three public LID implementations~\cite{Joulin2017, Kocmi2017, Duvenhage2017}. The LID implementation described in \cite{Joulin2017} is available on GitHub and is trained and tested according to a post~\footnote{\url{https://fasttext.cc/blog/2017/10/02/blog-post.html}} on the fasttext blog. Character (5-6)-gram features with 16 dimensional vectors worked the best. The implementation discussed in \cite{Kocmi2017} is available from \url{https://github.com/tomkocmi/LanideNN}. Following the instructions for an OSX pip install of an old r0.8~\footnote{\url{https://github.com/tensorflow/tensorflow/blob/r0.8/tensorflow/g3doc/get_started/os_setup.md}} release of TensorFlow, the LanideNN code could be executed in Python 3.7.4. Settings were left at their defaults and a learning rate of 0.001 was used followed by a refinement with learning rate of 0.0001. Only one code modification was applied  to return the results from a method that previously just printed to screen. The LID algorithm described in \cite{Duvenhage2017} is also available on GitHub.

The stacked classifier is also tested against the results reported for four other algorithms~\cite{Bestgen2017, Malmasi2015, Mathur2017, Medvedeva2017}. All the comparisons are done using the NCHLT~\cite{NCHLTTextCorpora2014}, DSL~2015~\cite{Zampieri2015} and DSL~2017~\cite{Zampieri2017} datasets discussed in Section~\ref{sec:related_works}.

\section{Results and Analysis}
\label{sec:results}

\begin{table}
  \caption{LID Accuracy Results. The models we executed ourselves are marked with~*. The results that are not available from our own tests or the literature are indicated with~'---'.}
  \label{tab:results}
  \centering
  \begin{tabular}{lllll}
    \toprule
    Model & Algorithm & NCHLT & DSL~'15 & DSL~'17 \\
    \midrule
    \midrule
    Joulin et al.~2017~\cite{Joulin2017}~*                & fasttext     & 93.30 & 93.20 & 88.60 \\
    Bestgen~2017~(DSL winner)~\cite{Bestgen2017}          & SVM          & ---   & ---   & 92.74 \\
    Medvedeva et al.~2017~\cite{Medvedeva2017}            & SVM          & ---   & ---   & 92.54 \\
    Malmasi \& Dras~2015~(DSL winner)~\cite{Malmasi2015}  & NB ensemble  & ---   & 95.54 & --- \\
    Mathur et al.~2017~\cite{Mathur2017}                  & RNN ensemble & ---   & 95.12 & --- \\
    Duvenhage et al.~2017~\cite{Duvenhage2017}~*          & NB+Lex       & 94.59 & ---   & --- \\
    Kocmi \& Bojar~2017~\cite{Kocmi2017}~*                & BRNN         & 67.84 & ---   & --- \\
    \midrule
    Naive-Bayes~only~*                                    & NB           & 94.36 & 94.98 & 91.89 \\
    Stacked~model~(NB)~*                                  & NB+NB        & 94.41 & 95.23 & 91.96 \\
    Stacked~model~(lexicon)~*                             & NB+Lex       & \textbf{96.12} & \textbf{99.34} & \textbf{98.70} \\
    Stacked~model~(50\% lex dropout)~*                    & NB+Lex       & 94.90 & 98.06 & 96.21 \\
    Lexicon~only~*                                        & Lex          & 82.88 & 97.86 & 93.56 \\
    Lexicon~only~(sans test data)~*                       & Lex          & 75.39 & 81.57 & 69.74 \\
    \bottomrule
  \end{tabular}
\end{table}

The average classification accuracy results are summarised in Table~\ref{tab:results}. The accuracies reported are for classifying a piece of text by its specific language label. Classifying text only by language group or family is a much easier task as reported in \cite{Duvenhage2017}. 

Different variations of the proposed classifier were evaluated. A single NB classifier (NB), a stack of two NB classifiers (NB+NB), a stack of a NB classifier and lexicon (NB+Lex) and a lexicon (Lex) by itself. A lexicon with a 50\% training token dropout is also listed to show the impact of the lexicon support on the accuracy.

From the results it seems that the DSL 2017 task might be harder than the DSL 2015 and NCHLT tasks. Also, the results for the implementation discussed in~\cite{Kocmi2017} might seem low, but the results reported in that paper is generated on longer pieces of text so lower scores on the shorter pieces of text derived from the NCHLT corpora is expected.

\begin{table}
  \caption{LID requests/sec. on NCHLT dataset. The models we executed ourselves are marked with~*. For the other models the results that are not available in the literature are indicated with~'---'.}
  \label{tab:perf_results}
  \centering
  \begin{tabular}{llllll}
    \toprule
    Model: & Joulin~2017~\cite{Joulin2017} & Duvenhage~2017~\cite{Duvenhage2017} & Kocmi~2017~\cite{Kocmi2017} & Proposed \\
    \midrule
    \midrule
    Performance: & 44k/s & 2.3/s & 0.75/s & 7.4/s \\
    \bottomrule
  \end{tabular}
\end{table}

The accuracy of the proposed algorithm seems to be dependent on the support of the lexicon. Without a good lexicon a non-stacked naive Bayesian classifier might even perform better.

The execution performance of some of the LID implementations are shown in Table~\ref{tab:perf_results}. Results were generated on an early 2015 13-inch Retina MacBook Pro with a 2.9 GHz CPU (Turbo Boosted to 3.4 GHz) and 8GB RAM. The C++ implementation in~\cite{Joulin2017} is the fastest. The implementation in~\cite{Duvenhage2017} makes use of un-hashed feature representations which causes it to be slower than the proposed sklearn implementation. The execution performance  of~\cite{Kocmi2017} might improve by a factor of five to ten when executed on a GPU.

\section{Conclusion}
\label{sec:conclusion}
LID of short texts, informal styles and similar languages remains a difficult problem which is actively being researched. The proposed algorithm was evaluated on three existing datasets and compared to the implementations of three public LID implementations as well as to reported results of four other algorithms. It performed well relative to the other methods beating their results. However, the performance is dependent on the support of the lexicon.

We would like to investigate the value of a lexicon in a production system and how to possibly maintain it using self-supervised learning. We are investigating the application of deeper language models some of which have been used in more recent DSL shared tasks. We would also like to investigate data augmentation strategies to reduce the amount of training data that is required.

Further research opportunities include data harvesting, building standardised datasets and shared tasks for South Africa as well as the rest of Africa. In general, the support for language codes that include more languages seems to be growing, discoverability of research is improving and paywalls seem to no longer be a big problem in getting access to published research.

%
%
%


\bibliographystyle{unsrt}
\bibliography{neurips_2019.bib}

%
%

\end{document}